\documentclass{article}
\usepackage{graphicx}
\usepackage{amssymb,amsfonts,amsmath}
\usepackage[numbers,sort&compress]{natbib}
\usepackage{multirow}
\usepackage{cjhebrew}
\usepackage{xcolor}
\usepackage{tipa}
\usepackage{authblk}
\usepackage[margin=1.25in]{geometry}
\definecolor{webgreen}{rgb}{0,.7,0}

\author[1]{Stephan C. Meylan}
\author[1]{Thomas L. Griffiths}
\affil[1]{University of California, Berkeley, Psychology}
\title{Word forms---not just their lengths---are optimized \\ for efficient communication}
\date{}

\begin{document}

\maketitle

\begin{abstract}
The inverse relationship between the length of a word and the frequency of its use, first identified by G.K. Zipf in 1935, is a classic empirical law that holds across a wide range of human languages. 
We demonstrate that length is one aspect of a much more general property of words: how distinctive they are with respect to other words in a language.
Distinctiveness plays a critical role in recognizing words in fluent speech, in that it reflects the strength of potential competitors when selecting the best candidate for an ambiguous signal.
\textit{Phonological information content}, a measure of a word's string probability under a statistical model of a language's sound or character sequences, concisely captures distinctiveness.
Examining large-scale corpora from 13 languages, we find that distinctiveness significantly outperforms word length as a predictor of frequency.
This finding provides evidence that listeners' processing constraints shape fine-grained aspects of word forms across languages. 
\end{abstract} 

\noindent Despite their apparent diversity, natural languages display striking structural regularities  \citep{greenberg1963, evansAndLevinson2009, futrellEtAll2015}. 
How such regularities relate to human cognition remains an open question with implications for linguistics, psychology, and neuroscience \citep{hauserEtAl2002, evansAndLevinson2009, kempRegier2012, fedzechkinaEtAl2012}.
Prominent among these regularities is the well-known relationship between word length and frequency: across languages, frequently-used words tend to be short \citep{zipf1935}.
In a classic work, Zipf \citep{zipf1935} posited that this pattern emerges from speakers minimizing total articulatory effort by using the shortest form for words that are used most often, following what he later called the \textit{Principle of Least Effort} \citep{zipf1949}. 
While the underlying cause has been the subject of debate \citep{yule1944, miller1957,ferrericanchoSole2003, conradMitzenmacher2004, piantadosi2014}, this relationship between word length and frequency remains one of the most robust statistical laws that describe human languages.


Here, we propose a generalization of Zipf's analysis and present two possible listener-focused explanations. 
We show that a word's frequency is inversely related to its  \textit{distinctiveness}---how easily it can be identified as the intended message for a given speech signal.
While speakers prefer easier-to-produce \textit{less} distinctive forms, they are constrained by listeners' need for \textit{sufficiently} distinctive forms to differentiate each word from others, especially if a speaker's intended word has higher-frequency competitors. 
If word recognition is modeled as Bayesian inference \citep{norrisMcQueen2008, ballingBaayen2012}, the probability of successful recognition depends on both the prior probability of the intended word and on the number and strength of alternative words (``competitors'').
We define a statistical measure of distinctiveness that succinctly captures the diagnosticity of a word form by assessing the aggregate strength of competitors in the language.
We then show that distinctiveness should be inversely related to frequency, if languages are constructed to equalize error rates for low and high frequency words.

Importantly, distinctiveness subsumes Zipf's observation regarding the relationship of length and frequency as a special case. Length is a na\"{i}ve approximation of the distinctiveness of a word form insofar as longer strings are simply less probable.
We demonstrate that a more comprehensive measure of distinctiveness that takes into account the sound-to-sound (phoneme-to-phoneme or letter-to-letter) sequences in a language accounts for significantly more frequency-related variance than does length across a broad sample of natural languages. 
This relationship between frequency and distinctiveness adds to a growing body of evidence that cognitive constraints influence the structural properties of natural languages \citep{hawkins1994, fedzechkinaEtAl2012, futrellEtAll2015}.

\section*{Model}
We define a probabilistic language model to characterize the distinctiveness of word forms in terms of their constituent sound-to-sound transitions.
This model can be fit using a large written sample of a particular language.
The starting point for the model is formalizing the task of the listener as a rational statistical inference.

\subsection*{Bayesian inference and distinctiveness}

Upon hearing a string of sounds $s$, a listener has to infer what word $w$ was intended by the speaker. This can be formulated as a problem of Bayesian inference. The listener should calculate a posterior distribution $P(w|s)$ over words based on the sounds. Applying Bayes' rule, this is given by
\begin{equation}
P(w|s) = \frac{P(s|w)P(w)}{P(s)}
\end{equation}
where $P(s|w)$ is the probability of hearing $s$ if $w$ is the intended word, $P(w)$ is the prior probability of the word $w$ intended by the speaker, and $P(s)$ is the probability of hearing $s$. 

Assuming that sounds are produced faithfully, such that $P(s_w|w)$ is close to $1$ for a particular string $s_w$ for each $w$ and close to $0$ otherwise, we obtain the approximation
\begin{equation}
P(w|s_w) \approx \frac{P(w)}{P(s_w)}
\label{eq:bayesapprox}
\end{equation}
which expresses the probability that the word $w$ is correctly identified as a function of its normalized frequency, $P(w)$, and the probability of the string $s_w$ in the language, $P(s_w)$. We define the \emph{distinctiveness} of a word to be inversely related to $P(s_w)$: intuitively, a word that shares the same sound sequences with many other words is necessarily less distinctive. 

Following this logic, we use the \textit{phonological information content} (PIC) of a word to measure the distinctiveness of its phoneme-to-phoneme transitions \citep{cohenPriva2008}, or its approximation in character-to-character transitions.
Analogous to the metric of lexical surprisal (negative log conditional probability) used to measure the predictability of a word given preceding words \citep{levy2008, piantadosiEtAl2011, smithLevy2013}, PIC is the negative log probability of the sequence of phonemes (distinct meaningful sounds) or letters in the string comprising a word.
To obtain a compact representation of the probabilities of various phone or character sequences in a language we estimate an \textit{n}-phone or \textit{n}-character model (analogous to an \textit{n}-gram model over words \citep{manningSchutze1999} but computed over individual phonemes or characters) from a corpus sample.
To support a stronger test of the relationship between distinctiveness and frequency, we avoid the circularity that more probable words necessarily contain more probable sequences by estimating the transition probabilities using the type inventory (unique words) in the language.
For the smaller datasets (those from the OPUS corpus) we also present the results for a model with token-weighted phonological transitions, though we note that interpreting the correlations obtained from the weighted model is challenging given this circularity.

Under this model, the phonological information content $PIC(w)$ of a word is defined as:
\begin{align}
\label{eq:PIC}  
PIC(w) &= -\log P(s_w) \\
&= - \log P(l_1,\ldots, l_{|s_w|}) \text{ for }l \in s_w 
\\
& = - \sum_{i = 1}^{|s_w|}{ \log P(l_i | l_{i -(n-1)}, \ldots, l_{i-1})}.
\end{align}
\vspace{-10px}

\noindent where $l$ are the phonemes or letters that comprise the sequence $s_w$, $|s_w|$ is the length of $s_w$, and \textit{n} is the order of the \textit{n}-phoneme or \textit{n}-character model (\textit{n}=5 in the current analyses to avoid overfitting).
PIC does not explicitly account for the morphemic components of a word (sub-word meaningful units, like the English prefix \textit{un-}), rather the the relative prevalence of morphemes is reflected in the character-to-character or phoneme-to-phoneme transition statistics.

\begin{table}[b]
\centering
\caption{\label{tab:motorcycleExample}Estimates of the information content of the word \textit{motorcycle} under uniform character probability (proportional to word length) and under a type-weighted model that takes into account sequential dependencies up to length 5 obtained from the 25,000 most frequent words in the English Google Books (2012) corpus. 
}
\setlength{\tabcolsep}{4pt} 
\renewcommand{\arraystretch}{1} 
\begin{tabular}{llllll}
``M''     & ``O''      & ``T''       & ``O''        & ... & $\Sigma$ Bits  \\
\hline   
\multicolumn{6}{l}{\textit{Uniform Character Probabilities}}    \\                 
$\log P(M)$  & $\log P(O)$   & $\log P(T)$    & $\log P(O)$     &     &       \\
4.700 & 4.700  & 4.700   & 4.700    & ... & 47.004 \\
\\
\multicolumn{6}{l}{\textit{5-Character Phonological Information Content Model}} \\                     
$\log P(M)$  & $\log P(O|M)$ & $\log P(T|MO)$ & $\log P(O|MOT)$ &     &       \\
5.358  & 3.223   & 3.451   & 6.110    & ... & 26.327 \\
\hline
\end{tabular}
\end{table}

\subsection*{Length and distinctiveness}
\begin{figure}[t]
\centering
\includegraphics[width=2.5in]{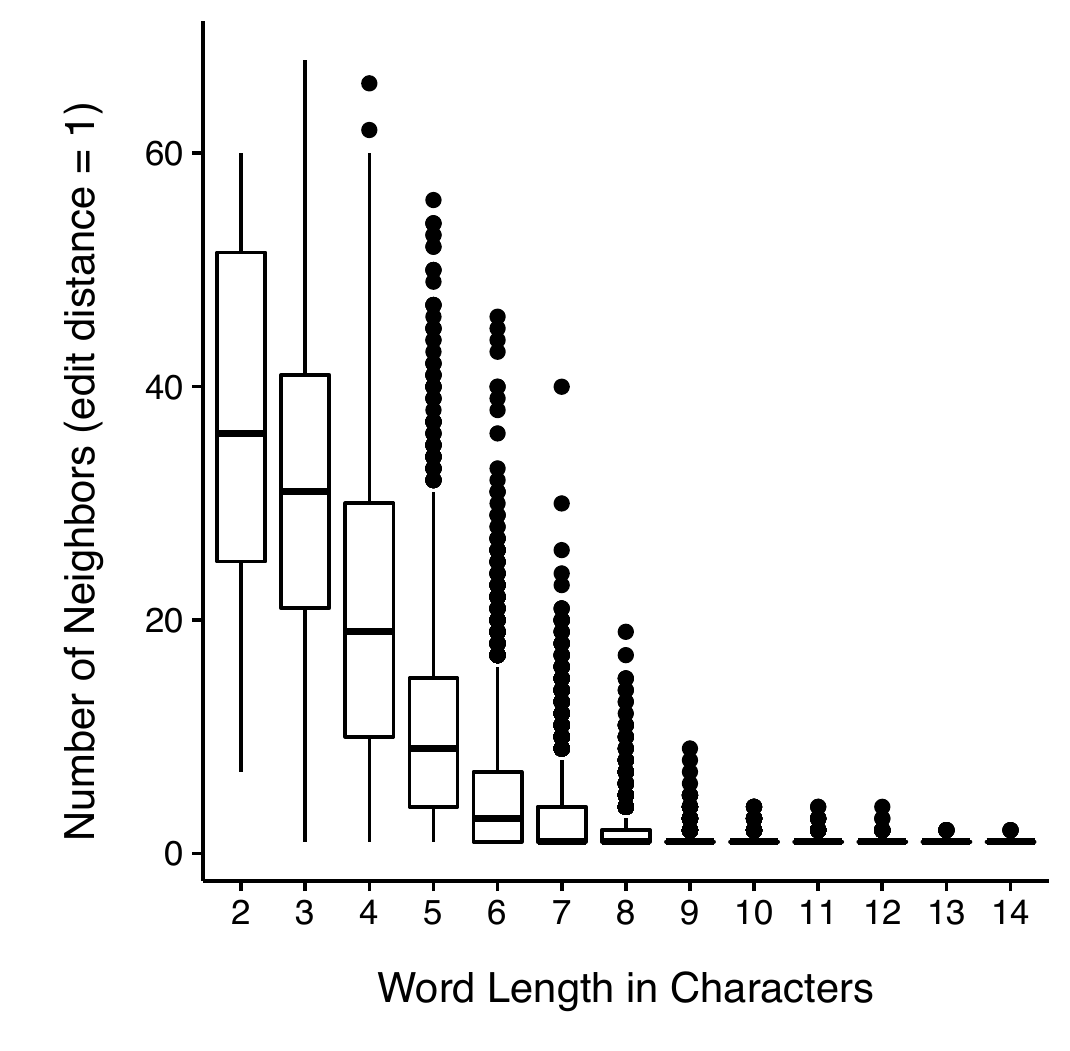}
\vspace{-5mm}
\caption{Longer words have fewer words with similar word forms (``neighbors''): the number of words within an edit distance of one phoneme decreases with the length of the word form. Data from 27,751 words in the English Clearpond dataset \citep{marianEtAl2012}.\vspace{-5mm}}
\label{neighborsPerLength} 
\end{figure}

Intuitively, short words have more similar competitors and are hence easier to confuse with other words, while long words have fewer neighbors (Fig.~\ref{neighborsPerLength}). 
Per Zipf's formulation, a signal which is too short may be ambiguous, and a listener is less likely to infer a speaker's intended meaning \citep{zipf1949}.
We argue that it is not length \textit{per se} which drives this effect, but rather the contribution of length to distinctiveness.

Word length is an important determinant of string probability: under a probabilistic treatment, a string that is one sound longer is a longer sequence of events, and hence of equal or lesser probability. In fact, length is strictly proportional to the probability (or log probability) of a string under a ``monkeys-on-typewriters'' model of letter-to-letter transitions such as that proposed by Mandelbrot \citep{mandelbrot1954SGS} and further explored by Miller \citep{miller1957}. 

Under this naive statistical model, symbols (sounds, or written characters approximating those sounds) are equiprobable and independent: $P(l_i) = 1/v$, where $v$ is the number of symbols in the symbol set of the language (its alphabet or phoneme inventory, for example 26 letters in English). 
If words were composed of equiprobable, independently-drawn symbols, $PIC(w)$ would simply be $|s_w| \log v$.

However, this approximation fails to capture regularities in the lexical substructure observed in natural languages in two obvious ways.
First, symbols are not equiprobable: across the word types in the English lexicon, \textit{w} is substantially less common than \textit{e}.
Second, the sound symbols are not statistically independent: the sound \textit{t} in English is followed more frequently by \textit{i} or \textit{e} and very rarely---if ever---by \textit{b} or \textit{g}.
People have rich knowledge of the relative prominence of these sequences in their respective languages---just as they have rich knowledge of inter-word statistical dependencies---and can call upon this knowledge in spoken word recognition \citep{vitevitchLuce1999, luceLarge2001}.
Ample psycholinguistic evidence also suggests that people are capable of using sub-word information, for example using sounds from the beginning of a word to predict possible continuations  \citep{marslenWilsonWelsh1978, marslen-Wilson1985, zwitserlood1989, eberhardEtAl1995}.

When the probability of sub-word sequences is taken into account, sequences of the same length can vary markedly in distinctiveness: while \textit{xylophone} and \textit{something} are both nine letters long, the latter is comprised of significantly less common subsequences. When phonological information content is computed with a more accurate model of the structure of words, it differs from length because it captures deviations from independence (see Table~\ref{tab:motorcycleExample}). 
 

\subsection*{Benefits of distinctiveness}
Conceiving of words in terms of their distinctiveness has several benefits compared to length.
First, distinctiveness (as measured by PIC) is a much more fine-grained measure of word form complexity, and  generates predictions contrary to word length in many cases. 
A shorter word may contain a relatively low probability phoneme sequence (e.g., \textit{depth}, 4 phonemes/5 letters, PIC = 17.86 bits under a type-weighted model from the Google Books 2012 English corpus), while a longer word may contain a higher probability, less informative sequence (e.g., \textit{ground}, 5 phonemes/6 letters, PIC = 12.85 bits under that same model).  

Second, PIC is closely related to metrics of lexical neighborhood density used in psycholinguistic models of spoken word recognition.
Neighborhood density reflects how many words have a similar form to a given word; while proposals vary on how to best measure neighborhood density, they share the intuition that words with more similar word forms (or ``neighbors'') are harder to recognize because there are more competitors consistent with a given signal.
PIC is formally very similar to  frequency-weighted neighborhood density \citep{lucePisoni1998}, however it measures the number of competitors at each successive phone---a feature consistent with empirical results suggesting incremental phoneme-by-phoneme processing in some cases \citep{eberhardEtAl1995}. 
PIC thus constitutes a more detailed measure of neighborhood density than the canonical measure of Coltheart's $N$ \citep{coltheartEtAl1977}, the number of words within a edit distance of one of the target word.
In particular, PIC is sensitive to competition effects from hearing the partial word form: while there are approximately 40 possible candidates upon hearing /\textipa{Ti}/ in \textit{thesis} in a large sample of English, neighborhood density as assessed by Coltheart's $N$ is much lower (the only competitors by this criterion are \textit{theses} and \textit{Theseus}).
We return to the question of whether treating spoken word recognition as a purely sequential prediction task is an appropriate simplifying assumption in the Discussion.

\subsection*{A relationship between distinctiveness and frequency}

We provide two listener-centric explanations as to why there might be an inverse relationship between distinctiveness and frequency.
First, distinctiveness provides a way to measure the effort that listeners expend in comprehending speakers; total comprehension effort is minimized when the most frequent words are the least distinctive. 
Zipf's original explanation for the inverse relationship between the frequency of a word and its length was based on the idea that languages are shaped by the desire of speakers to expend the least effort in producing words.
If longer strings are more effortful to produce, it makes sense that they should be associated with less frequent words.
However, we can imagine a similar argument being applied on the part of listeners: that languages are shaped by the desire of listeners to minimize the effort they expend in comprehending speakers.
Distinctiveness more accurately indexes comprehension effort, above and beyond word length.


Second, an inverse relationship between distinctiveness and frequency can also be derived from an invariance argument.
Recent work in computational psycholinguistics has discovered that a variety of syntactic phenomena can be predicted from the Uniform Information Density hypothesis: that languages are structured such that each word in an utterance provides approximately the same amount of information \citep{genzelCharnaik2002, aylettTurk2004, levyJaeger2007,piantadosiEtAl2011}. A similar invariance principle that might be relevant at the level of the words themselves is Uniform Recognizability: that the probability any word is successfully recognized is approximately the same. Under Equation~\ref{eq:bayesapprox}, the probability of successfully recognizing $w$ from its associated string $s_w$ is $P(w)/P(s_w)$. Making this constant across words means that we should expect $P(w)$ and $P(s_w)$ to be directly related, and hence $P(w)$ and $PIC(w)$ to be negatively correlated. 

To evaluate this hypothesis we evaluate the correlation between frequency and PIC across large corpus samples of 13 languages from three large-scale datasets: web scrapes from the Google 1T corpora \citep{googleEuropeanWeb1T, googleEnglishWeb1T}, scanned books from the Google Books 2012 corpus \citep{michelEtAl2011}, and subtitles from the 2013 OPUS corpus \citep{tiedemann2012}.
We limit our analysis to languages with phonemic scripts.
Following \citet{piantadosiEtAl2011}, we compute correlations over the 25,000 most frequent types in each language; following that same study we additionally compute the correlation between PIC and average in-context information content (trigram surprisal) over the same set of words.
\section*{Results}

We investigate the correlation between distinctiveness and frequency in large corpus samples (43m to 266b words) in 13 languages, across three large-scale datasets.
In each case we compute the frequency and in-context information content (mean trigram surprisal, or negative mean log conditional probability under a trigram model) for each word, and measure its PIC under an $n$-phoneme and $n$-character model estimated from distinct words in the language (for additional details, see Methods).
For the OPUS datasets, we also compute PIC under the analogous token-weighted models. 
If length is a reduced-resolution approximation of distinctiveness, then we expect an even stronger relationship between distinctiveness---as measured by PIC---and frequency across languages.

\subsection*{Cross-Linguistic Results}
\begin{figure}
\begin{center}
\includegraphics[width=3in]{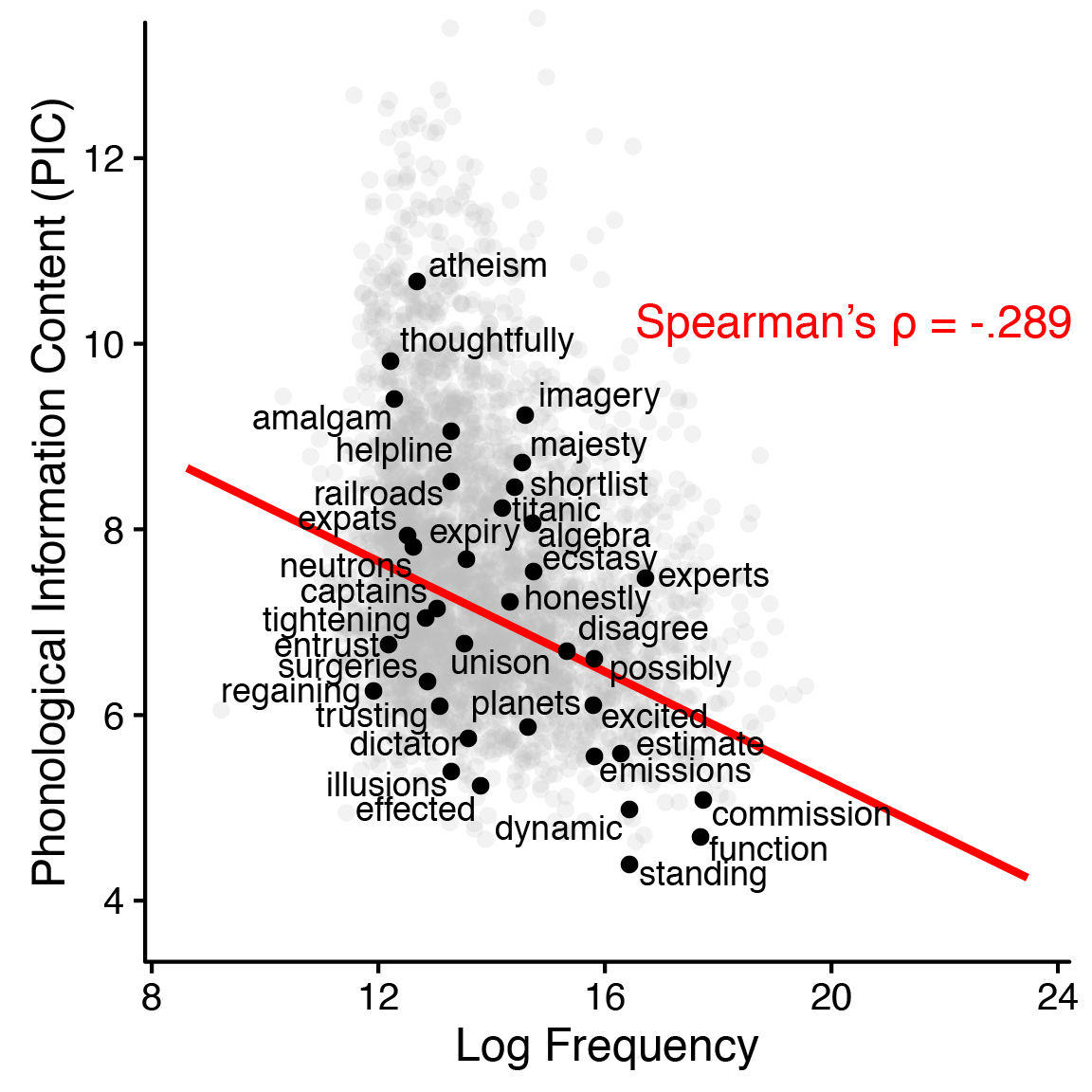}
\end{center}
\vspace{-5mm}
\caption{\label{fig:lengthConstant} 
For words with the same number of meaningful sounds (in this case 7 phonemes), distinctiveness measured with Phonological Information Content accounts for additional variance in frequency. The red line indicates the best linear fit for all words with seven sounds. Though data presented here reflect phonemic transcriptions, words are labeled in standard English form. Labeled points (black) are randomly sampled from the $n = 3,415$ words in the sample with phonemic transcription of this length.}
\end{figure}

\begin{figure*}
\begin{center}
\includegraphics[width=6in]{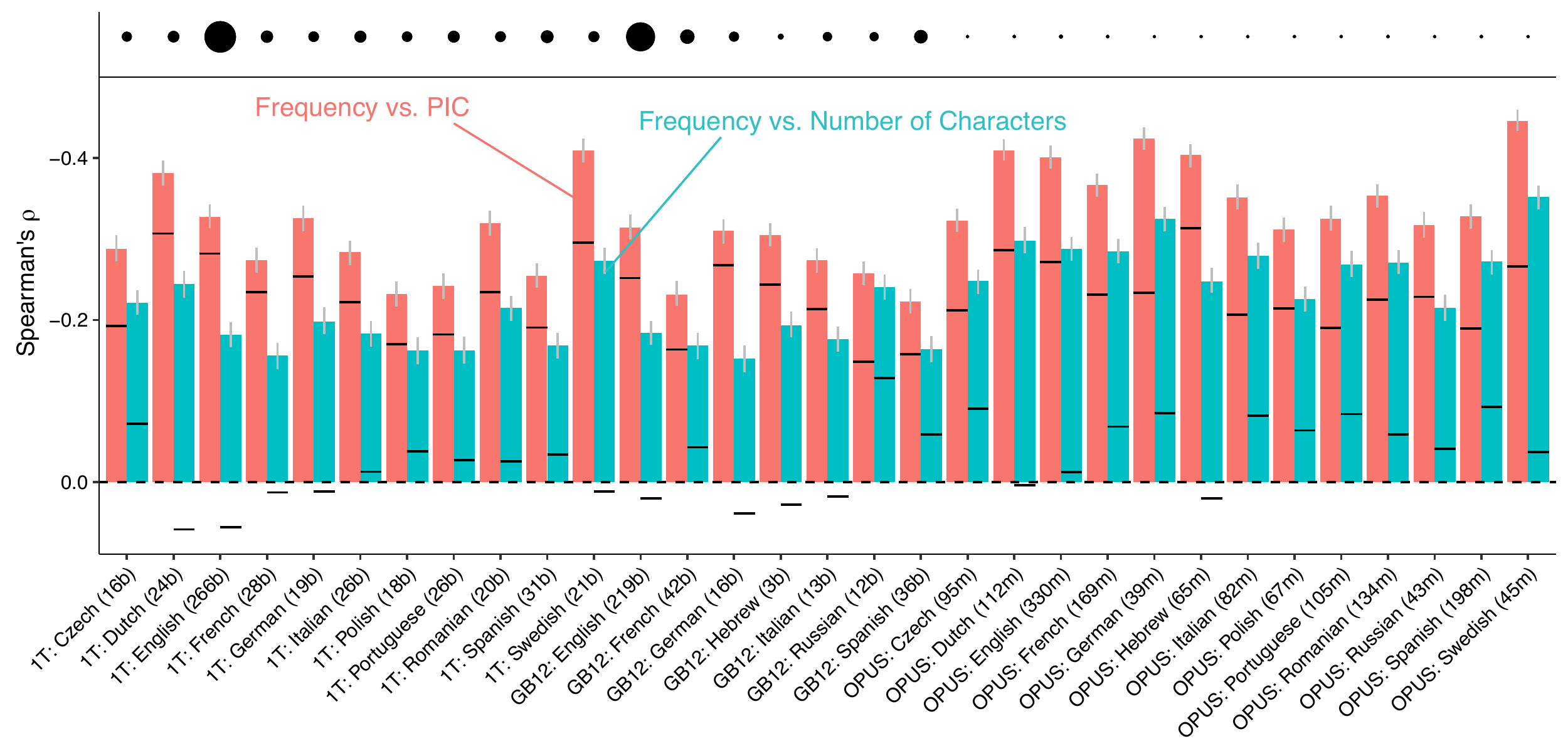}
\end{center}
\vspace{-5mm}
\caption{\label{dfss_unigram_sublex} 
Frequency demonstrates a stronger negative correlation with distinctiveness (as measured by type-weighted phonological information content, or PIC) in all datasets analyzed; this difference is significant in all cases but one (Russian in Books 2012). Bars indicate Spearman's $\rho$ for the two variables for the $n = 25000$ most frequent words in each dataset. Gray lines indicate the 99\% bootstrapped confidence interval. Black lines indicate the correlation with the other measure of word difficulty partialed out. After distinctiveness is partialed out, positive partial correlations are obtained for many datasets. Circles at the top of the figure indicate the number of tokens in each dataset. \vspace{-5mm}}
\end{figure*}

Following the methodology adopted in \citep{piantadosiEtAl2011}, we examine the correlation between word-level predictors (frequency and in-context information content) and metric of structural form (word length or PIC) for the 25,000 most frequent types in each language.
Unlike \citep{piantadosiEtAl2011}, we limit our analysis to in-dictionary types, thereby excluding person names, place names, and acronyms from the analysis.
We obtain a systematically stronger negative correlation between frequency and distinctiveness, as measured by type-weighted model PIC, than frequency and word length (Fig.~\ref{dfss_unigram_sublex}).
Even holding word length constant, distinctiveness explains substantial additional variance in word frequency (Fig. ~\ref{fig:lengthConstant}).
Building the model from phonemic transcriptions, this pattern holds in 11 of 11 languages in the Google 1T datasets, 6 of 7 languages from Google Books 2012, and all 13 languages from the 2013 OPUS corpus.
Building the model from raw characters---as an approximation of the phonological form---this pattern holds in all cases.
In many cases (10 of 11 languages in Google 1T, 3 of 7 in Google Books 2012, and 1 of 13 in OPUS) the partial correlation of PIC and frequency---with word length partialed out---is higher than the simple correlation of frequency and word length.
The obtained correlations are even stronger than those recently obtained between word length and average in-context information content \citep{piantadosiEtAl2011}.

\begin{figure*}[t]
\begin{center}
\includegraphics[width=5in]{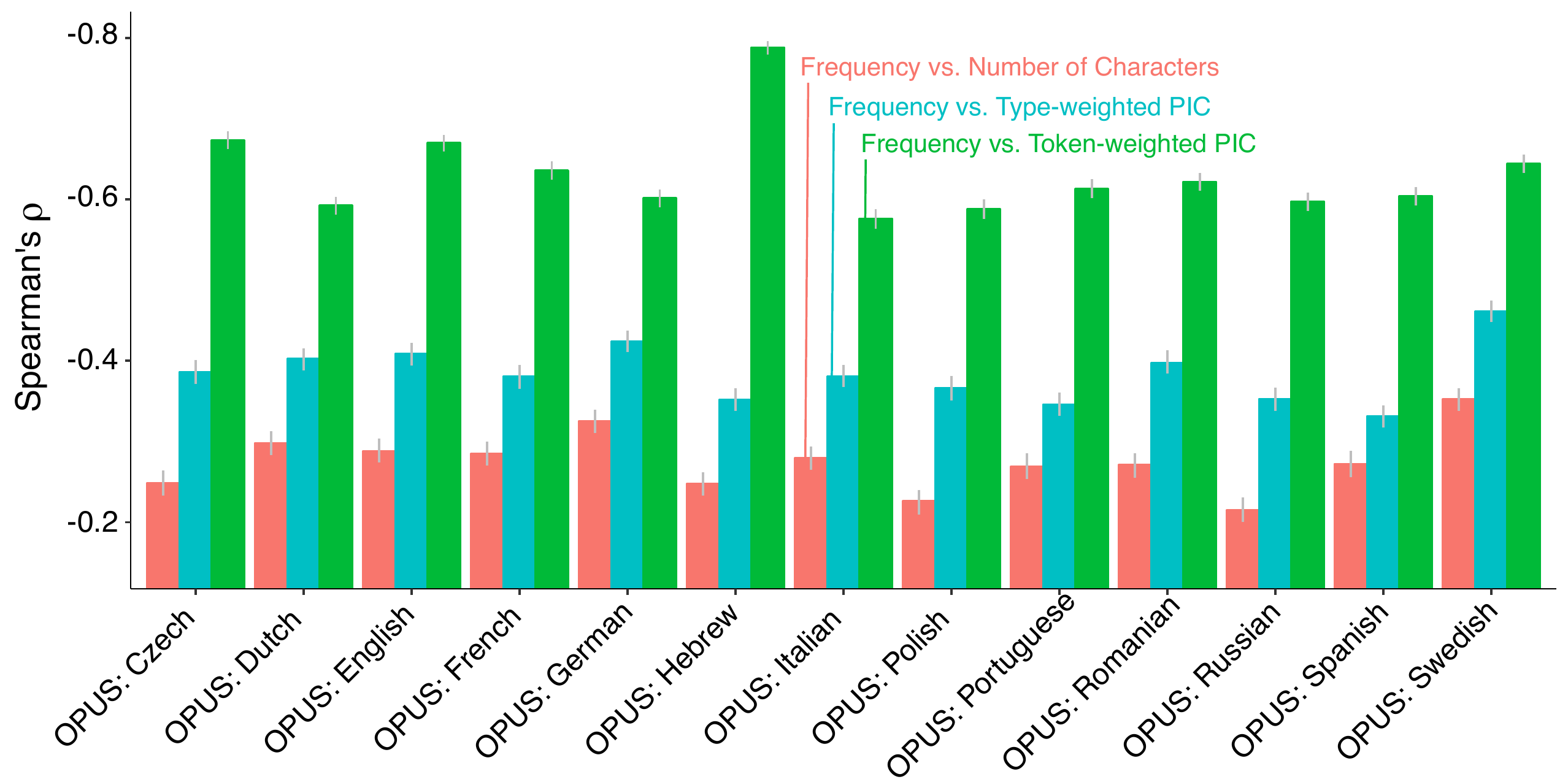}
\end{center}
\vspace{-5mm}
\caption{\label{token_character_sublex} 
Token-weighted phonological information content (PIC) is an even stronger predictor of frequency than type-weighted PIC. Bars indicate Spearman's $\rho$ for the two variables for the $n = 25000$ most frequent words in each dataset. Gray lines indicate the 99\% bootstrapped confidence interval in each case.}
\end{figure*}

PIC computed under the token-weighted model demonstrates an even stronger correlation with frequency across the languages in the sample (Fig.~\ref{token_character_sublex}).
We caution against an overly strong interpretation of this result, however, in that this correlation is not significantly higher than that obtained when phonemic content is shuffled among word forms (maintaining length for each word) and PIC subsequently recomputed.
In contrast, the correlation between frequency and PIC computed using a type-weighted model under the permuted dataset is no stronger than the correlation between frequency and length.

A similar pattern of results emerges regardless of whether the type-weighted model is computed over characters or phonemes; the sole exception is the Russian corpus from Google Books 2012 where word length is a stronger predictor when the model is computed over phonemes. 
However, this dataset is an outlier in two notable ways. 
Russian shows the lowest correlation between frequencies obtained from Google Books and OPUS (Pearson's $r$ = .48), as well as the lowest correlation between PIC estimates derived from Google Books and those derived from OPUS (Pearson's $r$ = .63).
Across languages, models built over phonemes and character transitions provide proportional estimates of PIC (Pearson's $r$ between .789 and .919 across languages, median = .874).
While more research is required to extend these findings beyond Germanic, Romance, and Slavic languages, Hebrew provides an important test of whether this relationship holds in languages with extensive nonconcatenative in addition to affixal morphology.

\subsection*{Relationship to Preceding Context}
\begin{figure}
\begin{center}
\includegraphics[width=3.35in]{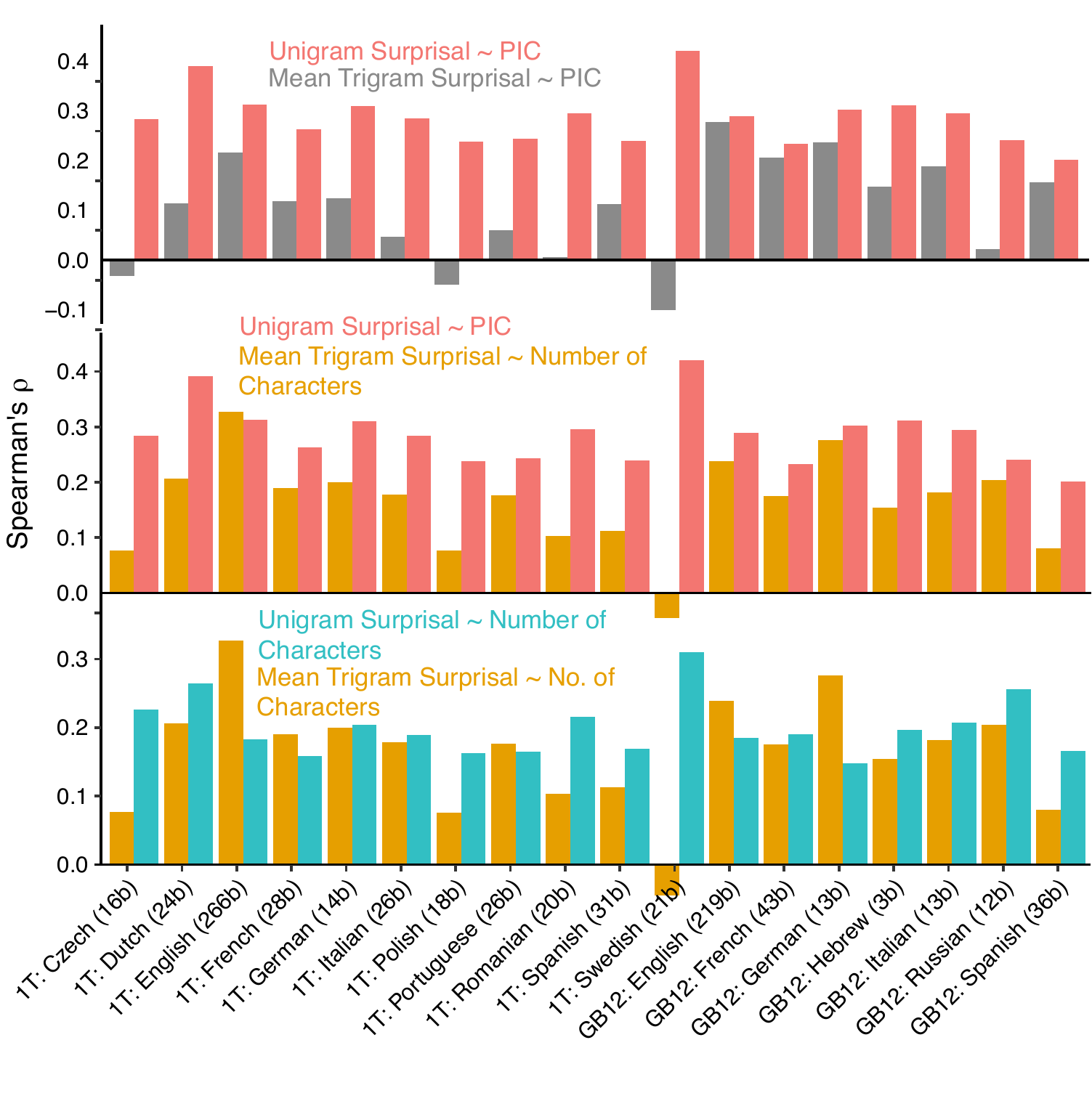}
\end{center}
\vspace{-10mm}
\caption{\label{figures/dfss_multiplot.pdf} 
Comparison of correlations between combinations of sentence-level predictability (unigram surprisal vs. mean trigram surprisal) and word-level measure (word length or phonological information content, PIC).  Bars indicate Spearman's $\rho$ for the two variables for the $n = 25,000$ most frequent words in each dataset.
\textbf{A}. PIC is more strongly correlated with unigram surprisal (negative log unigram probability) than mean trigram surprisal (negative mean log conditional probability) across a broad range of languages and datasets. 
\textbf{B.} The correlation between PIC and frequency is stronger than the correlation between mean trigram surprisal and length in all but one dataset (English Google 1T). The latter is the key relationship identified in \citet{piantadosiEtAl2011}. 
\textbf{C.} Among words found in a relevant language-specific dictionary, unigram surprisal (negative log normalized frequency) is a better predictor than mean trigram surprisal of word length in most cases. Notable exceptions are English and French in Google 1T and English and German in Google Books 2012.
Correlations from the OPUS datasets are omitted because these datasets are too small to reliably estimate mean trigram surprisal for most languages.
 \vspace{-5mm}}
\end{figure}

\textbf{}
While \citet{piantadosiEtAl2011} found that taking into account contextual predictability / information content (in the form of mean trigram surprisal in a dataset) better predicts word length than using a simple frequency measure (operationalized as unigram surprisal, or negative log probability), we find a qualitatively different pattern of results for PIC.
In the analysis above, we find that the correlation between frequency and PIC is \textit{higher} than the correlation between mean trigram surprisal and PIC in all cases (Fig. \ref{figures/dfss_multiplot.pdf}, A).
The correlation between frequency (negative log unigram probability) and PIC is also greater than the correlation between mean trigram surprisal and word length in all but one case (Fig. \ref{figures/dfss_multiplot.pdf}, B). 
We find substantially attenuated support for the principal claim in \citet{piantadosiEtAl2011}, in that we find unigram frequencies better predict word length than does mean trigram surprisal (Fig. \ref{figures/dfss_multiplot.pdf}, C).
This discrepancy, which we explore further in the Discussion, seems to reflect refinements in the list of lexical items analyzed, improvements in the data preparation and analysis methodology in the current work, and underlying issues in the construct validity of average information content for word forms in languages with rich morphology.

\section*{Discussion}
The relationship between word length and frequency is one of the most robust empirical findings regarding the structure of natural languages.
Treating word recognition in terms of Bayesian inference, we show that the requirement for successful recognition not only motivates the relationship between frequency and length, but generates even stronger predictions regarding the relationship between frequency and the probability of the phoneme sequences (and their approximation in terms of character sequences) that make up word forms.
Analyses of large-scale corpora from 13 languages across three datasets (including the linguistic content of web pages, books, and movie subtitles) substantiate these claims.
In this section we consider some of the implications of these results, as well as caveats and future directions for research.

\begin{figure}
\begin{center}
\includegraphics[width=3in]{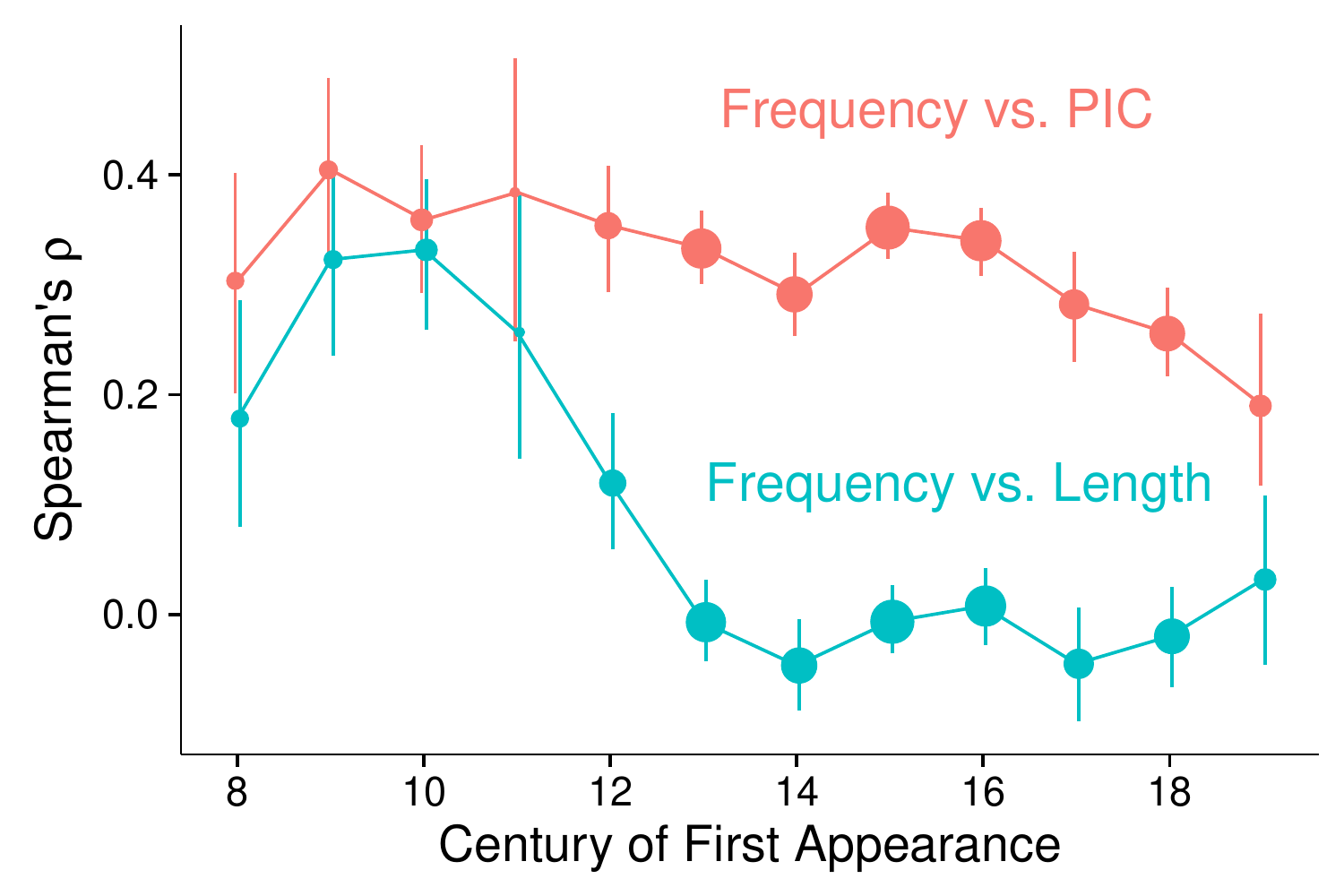}
\end{center}
\vspace{-5mm}
\caption{\label{figures/histPlot.pdf} 
For words that have entered English more recently, the correlation between frequency and PIC is higher than the correlation between frequency and word length. This is consistent with the hypothesis that PIC can capture finer-grained changes in word forms---which are more likely on shorter timelines--besides changes in length. Error bars show bootstrapped 99\% confidence intervals.\vspace{-5mm}
}
\end{figure}


The approach that we have taken in this paper potentially provides  a link between the psycholinguistic processes at work in producing and perceiving speech and the factors that shape human languages. Previous research has demonstrated that speakers may ``reduce''---underarticulate, shorten, weaken, or omit altogether---frequent or highly predictable words in specific contexts, for example using ``probly'' in place of ``probably'' in fluent speech \citep{aylett2006, bellEtAl2009, gahlEtAl2012}. 
On longer timescales, variants with more probable forms  may become the dominant form in the language, e.g., English \textit{lunch} displaced \textit{luncheon} in the 19\textsuperscript{th} and 20\textsuperscript{th} centuries \citep{oed2016} (see \citet{mahowaldEtAl2013} for additional examples). 

PIC provides a metric by which phonemic and phonetic lexical variants may be compared while capturing variation beyond that of word length.
Whereas word length can only reflect deletion and epenthesis (the insertion of phonemic material), PIC is sensitive to changes that maintain the same length, including assimilation (a sound becoming more similar to a neighboring sound), dissimilation (a sound becoming less similar to a neighboring sound), lenition (a consonant taking on more vowel-like qualities), and metathesis (transposition of proximal sounds).
The change from Middle English \textit{aks} to \textit{ask}, for example, is reflected in a change in PIC, though both have the same length. 
Paradigmatic sound changes operating across many words---or indeed an entire language (e.g. phonemic mergers)---can also be characterized in terms of changes in PIC, similar to entropy-based estimates of functional load \citep{surendran2006quantifying, wedel2013high}.

The formulation of the relationship between word form distinctiveness and frequency that we present above also makes several testable predictions regarding the relationship between frequency, length, PIC, and how long a word has been in a language.
First, we expect words that have recently entered a language such as neologisms and loanwords from other languages to exhibit a relatively weak relationship between frequency and their structural form, regardless of whether that structural form is characterized in terms of either PIC or length. 
In the case of neologisms, semantic transparency of the inputs initially determines structural form; in the case of loanwords, the it is heavily influenced by the phonological form in the language from which the word is borrowed.
Second, we expect a word's structural form comes to better reflect frequency over the course of time, in that changes that improve communicative efficiency can proceed gradually, in part because variation is constrained by the need to maintain intelligibility within a population of speakers.
Third, frequency should demonstrate a stronger correlation with PIC than with length on shorter timescales, in that PIC is sensitive to more subtle changes in the word form than the change in length. 
For example, deletion of a phoneme often comes only after an intermediate stage where that phoneme is weakened; PIC quantitatively reflects the intermediate weakening, whereas word length changes when the phoneme is deleted entirely.

In order to test these hypotheses, we obtain the date of first appearance for 31,027 English types from the Oxford English Dictionary.
We stratify these dates by century, assigning pre-700 (corresponding to proto-Old English) words to the 7th century, and compute Spearman's rank correlation coefficient against current word frequency for the set of words corresponding to each century.
The correlations we obtain between PIC and frequency are stronger than those between length and frequency for words entering English between the 13th and 20th century  (Fig. ~\ref{figures/histPlot.pdf}).
This pattern of results provides qualitative support for all three hypotheses: an overall weaker relationship between frequency and structural form among recent words, an increasingly strong relationship for words that have been in the language for a longer interval, and a stronger relationship between PIC and frequency for those words that have entered the language more recently.
With additional datasets we hope to characterize how changes in frequency relate to specific changes in word form.

In contrast to the original formulation by Zipf which focused on minimization of speaker effort, in-context predictability as presented in \citet{piantadosiEtAl2011} and word form distinctiveness both implicate listener-oriented pressures in the relationship between frequency and word form.
The two proposals make contrasting predictions regarding what factors are most relevant to generalizing the relationship, however. 
In-context predictability generalizes the notion of a listener's prior beliefs regarding the probability of a given word, replacing frequency with predictability.
Distinctiveness provides a more detailed characterization of word form complexity.     
In principle, both proposals could be true: the strongest relationship could be between in-context predictability and word form distinctiveness, though we do not find empirical support for this in the datasets analyzed here.
Instead, as noted in the results we find in the current work substantively attenuated support for the relationship between word length and average in-context predictability (Fig. \ref{figures/dfss_multiplot.pdf}). 
Here we further investigate the source and significance of this discrepancy with the findings of \citet{piantadosiEtAl2011}.

\begin{figure*}
\begin{center}
\includegraphics[width=6in]{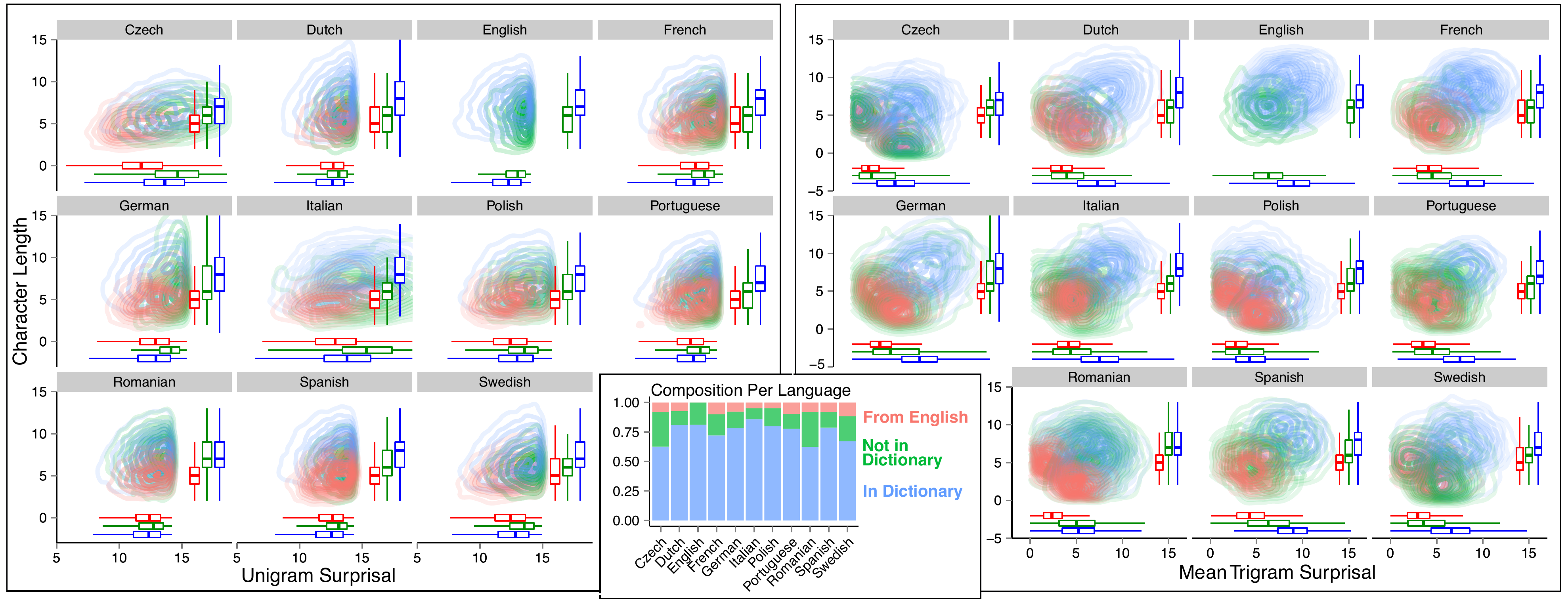}
\end{center}
\caption{\label{bothPredictors_marginals} 
2-D density plots depicting the relationship between sentence-level predictor (unigram surprisal or mean trigram surprisal) and word length for the 25,000 most frequent words in 11 languages used in \citet{piantadosiEtAl2011}. Words are stratified into three categories: those in a spelling dictionary of the target language (blue), those in a spelling dictionary for English (red), and those not in either (typically names and abbreviations, green). High correlations between mean trigram surprisal and word length, as well as low correlations between unigram surprisal emerge from an interaction between these groups.  Marginal boxplots show the median, inter-quartile range (IQR), and 1.5 * IQR for each of the groups. Densities are normalized per group.}
\end{figure*}

Using the same list of words and in-context information content estimates from \citet{piantadosiEtAl2011}, we classify each word into one of the three categories: those found in the relevant dictionary (by testing for membership in the corresponding GNU Aspell dictionary for each language), those found in English (by testing for membership in the English Aspell dictionary for words in non-English languages), and label the remainder as out-of-dictionary.
We designate those words found both in the language-specific dictionary as well as English, e.g. Spanish \textit{pan}, as in-dictionary items. 
The proportion of items found in each category for each language, as well as example classifications from Spanish, are presented in the inset in Fig. \ref{bothPredictors_marginals}, center.

\begin{figure*}
\begin{center}
\includegraphics[width=6in]{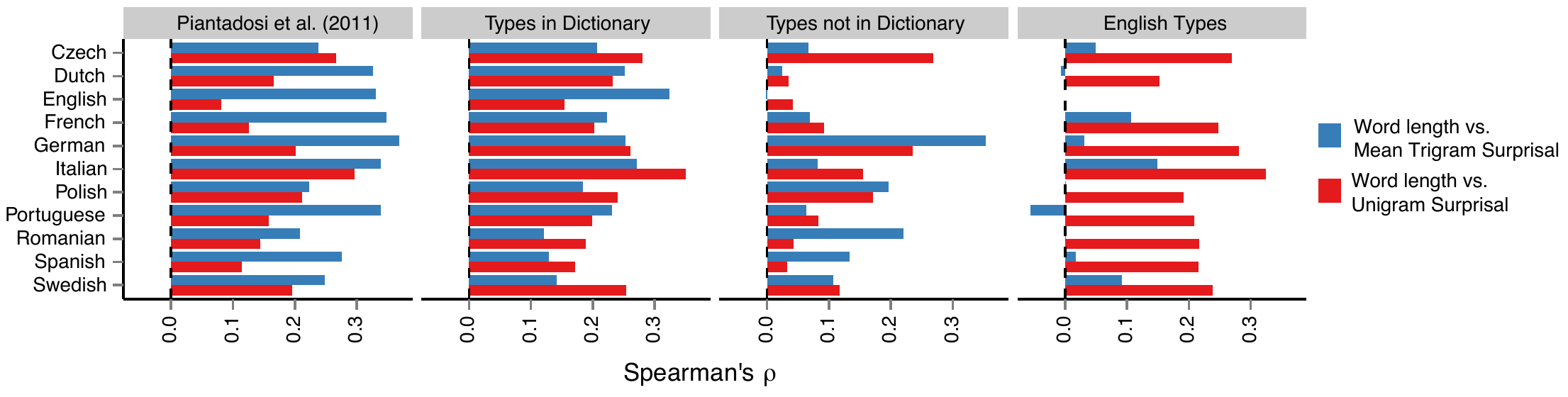}
\end{center}
\caption{\label{4groups_replicationPlot} 
\citet{piantadosiEtAl2011} found higher global correlations between word length and in-context predictability (information content) as measured by mean trigram surprisal (blue bars in panel 1) than between word length and frequency (unigram surprisal), red bars in panel 1). The correlation is substantially weaker within each of three sub-groups of lexical items: types that are found in the relevant dictionary (panel 2), those found in English (panel 4), and those found in neither dictionary (panel 3).}
\end{figure*}

Under this analysis, within-group correlations for all three groups are substantially lower than the aggregate correlations reported in \citet{piantadosiEtAl2011} (Fig. \ref{4groups_replicationPlot}).
Instead, the high global correlation between mean in-context information content and word length emerges from the inclusion of out-of-dictionary items and (for languages other than English) items from English, which are both more predictable and shorter than in-dictionary words (Figure \ref{bothPredictors_marginals}, right).
The global correlations between frequency and word length found in that study were depressed by out-of-dictionary items, which were shorter---yet oddly \textit{less} frequent---than in-dictionary terms (Figure \ref{bothPredictors_marginals}, left).  

Because \citep{piantadosiEtAl2011} represented word forms as closest ASCII equivalents (e.g., \textit{manana} for Spanish \textit{ma\~{n}ana}), some words cannot be found in the relevant dictionary.  
This means that for the above analysis some lexical items that should have been classified as in-dictionary have been classified as out-of-dictionary, with unknown implications for the within-group correlations.
Because there is no way to systematically restore these word forms, we compare the correlation between in-context information content (mean trigram surprisal, which they also refer to as \textit{information content}) and word length, as well as unigram surprisal (negative log normalized frequency) and word length using estimates from the main analysis.

Using 25,000 in-dictionary words---and respecting the original character encoding---we find that frequency is a better predictor of word length than in-context information content in most languages among those words that appear in the corresponding language's dictionary. 
Key exceptions, however, are consistent with the findings of \citet{piantadosiEtAl2011}: English and French in the Google 1T and English and German in Google Books 2012 show a stronger correlation of mean trigram surprisal and word length (Fig. \ref{figures/dfss_multiplot.pdf}, C).
This may reflect that strong correlations only emerge with unbiased estimates of in-context information content, which may in turn only be obtained in very large datasets: English Google 1T and Books 2012 corpora are at least five times larger than the next largest dataset.
In the case of Google 1T, this correlation is higher than that of frequency and PIC, leaving open the possibility that word length better reflects in-context information content while PIC more strongly reflects frequency.

While the heterogeneous pattern of results in which structural form is more closely linked to frequency in some languages and to in-context information content in others may reflect genuine cross-linguistic differences, in the current section we argue that computing in-context information content using $n$-gram models may be theoretically unsound in morphologically-rich languages. 
Consider for example that whereas an $n$-gram model for English would have entries for a handful of forms for the verb \textit{sell} (e.g., \textit{sell, sells, sold, selling}), an $n$-gram model for Spanish, a language with much richer morophology for verbs, needs to have many more entries for the corresponding verb \textit{vender} owing to the combinatorial space of possible conjugations and object clitics ($\approx$160 in Google Books 2012). 
Aggregating across inflected forms, the forms of \textit{vender} and \textit{sell} are approximately equally common, yet estimates of mean in-context surprisal may are more likely to be biased in Spanish given the sparsity of certain forms (e.g., \textit{v\'{e}ndeselas}, corresponding to English imperative \textit{you sell them that}).
Second, if an analysis is conducted on the top $n$ most frequent word forms, high frequency lemmas take up more of the list because they are spread across a larger number of inflected entries.
Depending on what parts of speech have high morphological complexity, substantive differences may emerge in the composition of the vocabulary under analysis across languages.
Future work will need to investigate how morphological complexity interacts with in-context information content across languages, and whether different tokenization (segmentation) processes should be used, for example determining whether object clitics should be treated as separate words in Spanish.
For now, we note that we obtain the same pattern of results for PIC and frequency when we restrict our analysis to morphologically simple word forms for those languages with morphological annotation in the CELEX database \citep{baayenEtAl1995} (English, German, and Dutch).

We make two key simplifications in the current work with respect to the measurement of PIC: we use a relatively simple, purely sequential representation of the phonemes that comprise a word and we compute PIC with respect to the true preceding context within each word. 
Using a richer hierarchical representation of word form structure may account for regularities than an $n$-phoneme model cannot, leading to a modest improvement in predicting a held-out set of words \citep{futrellInPress}.
Treating spoken word recognition as sequential recognition of phonemes may also discount regularities in morophological structure, though in many cases such regularities are captured implicitly in the transition probabilities among phonemes. 
Adapting more elaborate models of lexical structure---both morphological and phonological---may lead to improved estimates of word form probability in a language.

With respect to the second simplification, listeners do not have access to the true identity of preceding phonemes, and instead likely marginalize over a distribution of preceding as well as following sounds within a word in the process of word recognition \citep{strand2016}.
This may mean that listeners have comparatively peaked estimates of the preceding sounds only in the case of longer words,, such that the relationship between frequency and PIC as computed here is weaker for short words.
In future work we intend to investigate whether a measure of uncertainty that takes this into account can explain additional variance in word frequency. 

\section*{Data Issues}
Corpus analyses of the scale used here necessarily contain some degree of noisy data.
The choice of texts, methods for identifying content (i.e., excluding page numbers, tables, publisher notes etc.), optical character recognition procedure, and tokenization all influence the obtained dataset.
Additional data preparation decisions such as text encoding, treatment of punctuation, and treatment of upper and lower case forms can all have pronounced effects on lexical statistics.

Despite these challenges, these datasets constitute the best resource available for computing average in-context information content for lower frequency words, given that their huge size provides better evidence of the range of contexts in which these words appear. 
In the course of the analysis we found two notable issues in both the Google Books 2012 and Google 1T datasets;  we include a brief note on each here to alert others using this or similar corpora.

First, we found extensive evidence of issues with word segmentation in the Hebrew corpus of Google Books 2012 (for details of the rule-based tokenization procedure see \citet{lin2012syntactic}). 
We find a high proportion of bound morphemes  (\begin{cjhebrew}l/s\end{cjhebrew}, \begin{cjhebrew}t'\end{cjhebrew}, \begin{cjhebrew}l\end{cjhebrew}, \begin{cjhebrew}/s\end{cjhebrew}, \begin{cjhebrew}b\end{cjhebrew}, \begin{cjhebrew}w\end{cjhebrew}, \begin{cjhebrew}h\end{cjhebrew}) listed as separate lexical items in the raw data, in total accounting for approximately 20\% of the total unigram probability mass.

Second, the web-based nature of the Google 1T corpora strongly influences certain average in-context surprisal estimates. 
For example, ``Romanian'' has very low trigram surprisal in the Google 1T corpus because it is overwhelmingly encountered in language lists or lists of currencies (22\% of instances appear in the context ``Polish Portuguese Romanian''). 
Furthermore, the statistical properties of the everyday language environment of speakers may differ substantially from the corpora examined here.
Nonetheless, these corpora constitute the best approximation to naturalistic language use of sufficient size to support lexicon-wide analysis, particularly the calculation of mean information content for relatively rare words. 
As more corpora are made available, we hope to extend the analyses presented above to a more diverse set of languages.
\section*{Conclusion}
The canonical inverse relationship between word length and frequency is a special case of an even broader relationship between word distinctiveness and frequency.
Distinctiveness plays a crucial role in word recognition, capturing the strength of competing targets for a speech signal. 
Rational analysis reveals limitations on word forms: speakers can simplify and shorten words, but they are limited by listener's requirements for distinctive word forms for successful recognition.
\section*{Methods}

\subsection*{Datasets for Frequency and Surprisal Estimates}
The Google Web 1T datasets were downloaded from the Linguistic Data Consortium \citep{googleEnglishWeb1T, googleEuropeanWeb1T}; the Google Books 2012 datasets were downloaded from \textbf{storage.googleapis.com/books/ngrams/books/ datasetsv2.html} \citep{michelEtAl2011}, and OPUS (2013) from \textbf{opensubtitles.org}  \citep{tiedemann2012}.
All $n$-grams with punctuation-only words were discarded, and punctuation appearing with other text, with the exception of apostrophes, was removed. 
We make the simplifying assumption that the tokenized orthographic forms correspond to psychologically salient words in the lexicon of speakers; while this assumption does not hold for all forms (e.g. German compound nouns), it holds for the vast majority of word forms in the analysis (see also our analysis of morphologically-simple forms in the Discussion, which also addresses this question).
All characters were converted to lowercase using the relevant POSIX locale; US English and European Portuguese were used for English and Portuguese, respectively.
In the case of Google Books 2012, records with part-of-speech tags were discarded, along with records from earlier than 1800.
UTF-8 encoding was maintained throughout for all languages and datasets.
Hebrew strings were represented with right-normalized forms.
Counts were stored using ZS, a specialized file format for efficient retrieval of $n$-gram counts \citep{smithSubmitted}.

\subsection*{Estimating Sentential Information Content}
Following Piantadosi et al. (2013) we analyze a word list constructed from the 25,000 most frequent words in each dataset.
Token frequencies were computed from the 2013 release of the the OPUS subtitle corpus.  
For each language, the list of unique types were filtered by those words recognized by the UNIX utility Aspell for the relevant locale.
After filtering, we computed the negative log unigram probability (proportional to log normalized frequency) for each word $w$ along with the negative mean log trigram probability across contexts, following \citep{piantadosiEtAl2011},  $- \frac{1}{N} \sum^{N}_{i=1}{\log P(W = w| C = c_i)},$ where $c_i$ is the context for the $i$th occurrence of $w$ and $N$ is the frequency of $w$ in the dataset.

\subsection*{Estimating Phonological Information Content}
For the type-weighted models, a five-character transition model was estimated for each language using the 25,000 most frequent in-dictionary words also appearing in the corresponding OPUS subtitle corpus.
For the token-weighted models (OPUS only), a five-character transition model was estimated using all in-dictionary tokens in the corresponding OPUS subtitle corpus.
In both the token- and type- weighted cases, we also produced a five-phone transition models for all languages with the exception of Hebrew using IPA transcriptions from an automatic speech synthesizer, eSpeak. 
Using IPA representations for words accounts for language-specific variations in orthographic conventions.
For example, written Spanish includes accents only when the placement of prosodic stress cannot be deduced from more general rules in the language.  
Using an IPA transcription avoids the need for developing language-specific decisions, for example deciding whether `a' vs. `\'{a}' should be merged or kept as separate orthographic variants in Spanish. 
Loan words and acronyms can greatly affect the obtained transition probabilities, especially when the transitions observed in the type-weighted model (e.g., if the transitions in ``Okeechobee,'' ``ma\~{n}ana,'' and ``ACLU'' are as heavily weighted in a phonotactic model of English as the transitions in ``they'' and ``will''). 
To minimize these effects, we use only non-capitalized types present in Aspell dictionaries to build sound and character transition models for each language (with the exception of German, in which common nouns are capitalized).
To avoid overfitting among higher order sequences, phone and character transition probabilities for the type-weighted models were computed with modified Kneser-Ney smoothing \citep{chenGoodman1999} with interpolation on orders 3, 4, and 5 using the SRILM  toolkit \citep{stolcke2002}. 
Good-Turing smoothing was used for the token-weighted models.
Each word's phonological probability was calculated as the product of the probabilities of each symbol given the preceding symbol string, including a start symbol $\star$ , e.g., $P(the) = P(t|\star)  \times P(h|\star t) \times P(e|\star th)$.
The probability of the end symbol was omitted in that this appreciably inflates the PIC of short words under a type-weighted model (``to'' is an unlikely/high-surprisal sequence because there are many words that begin with ``to'').

\subsection*{Code availability}
Our library for the rapid cleaning, manipulation, summarization, and querying of $n$-gram data is available at \textbf{github.com/smeylan/ngrawk}.
Jupyter notebooks for the analyses presented here are available at \textbf{github.com/smeylan/pic-analysis}.
\section*{Acknowledgements}
This material is based upon work supported by the US National Science Foundation Graduate Research Fellowship under grant no. DGE-1106400 and NSF grant no. SMA-1228541. Special thanks to Steven Piantadosi for sharing materials and lexical information content estimates, helpful commentary on early drafts from Terry Regier, and members of the Computational Cognitive Science Lab at UC Berkeley for valuable discussion.

{\scriptsize
\bibliographystyle{unsrtnat}
\bibliography{sublexical_bibliography}}

\begin{thebibliography}{52}
\providecommand{\natexlab}[1]{#1}
\providecommand{\url}[1]{\texttt{#1}}
\expandafter\ifx\csname urlstyle\endcsname\relax
  \providecommand{\doi}[1]{doi: #1}\else
  \providecommand{\doi}{doi: \begingroup \urlstyle{rm}\Url}\fi

\bibitem[Greenberg(1963)]{greenberg1963}
JH~Greenberg.
\newblock Some universals of grammar with particular reference to the order of
  meaningful elements.
\newblock In JH~Greenberg, editor, \emph{Universals of Human Language}, pages
  73--113. MIT Press, Cambridge, MA, 1963.

\bibitem[Evans and Levinson(2009)]{evansAndLevinson2009}
N~Evans and SC~Levinson.
\newblock {{T}he myth of language universals: language diversity and its
  importance for cognitive science}.
\newblock \emph{Behav Brain Sci}, 32\penalty0 (5):\penalty0 429--448, 2009.

\bibitem[Futrell et~al.(2015)Futrell, Mahowald, and Gibson]{futrellEtAll2015}
R~Futrell, K~Mahowald, and E~Gibson.
\newblock {{L}arge-scale evidence of dependency length minimization in 37
  languages}.
\newblock \emph{Proc. Natl. Acad. Sci. U.S.A.}, 112\penalty0 (33):\penalty0
  10336--10341, 2015.

\bibitem[Hauser et~al.(2002)Hauser, Chomsky, and Fitch]{hauserEtAl2002}
MD~Hauser, N~Chomsky, and WT~Fitch.
\newblock {{T}he faculty of language: what is it, who has it, and how did it
  evolve?}
\newblock \emph{Science}, 298\penalty0 (5598):\penalty0 1569--1579, 2002.

\bibitem[Kemp and Regier(2012)]{kempRegier2012}
C~Kemp and T~Regier.
\newblock {{K}inship categories across languages reflect general communicative
  principles}.
\newblock \emph{Science}, 336\penalty0 (6084):\penalty0 1049--1054, 2012.

\bibitem[Fedzechkina et~al.(2012)Fedzechkina, Jaeger, and
  Newport]{fedzechkinaEtAl2012}
M~Fedzechkina, TF~Jaeger, and EL~Newport.
\newblock {{L}anguage learners restructure their input to facilitate efficient
  communication}.
\newblock \emph{Proc. Natl. Acad. Sci. U.S.A.}, 109\penalty0 (44):\penalty0
  17897--17902, 2012.

\bibitem[Zipf(1935)]{zipf1935}
GK~Zipf.
\newblock \emph{The Psychobiology of Language}.
\newblock Houghton-Mifflin, 1935.

\bibitem[Zipf(1949)]{zipf1949}
G~Zipf.
\newblock \emph{Human {B}ehaviour and the {P}rinciple of {L}east-{E}ffort}.
\newblock Addison-Wesley, Cambridge, MA, 1949.

\bibitem[Yule(1944)]{yule1944}
G.U. Yule.
\newblock \emph{The Statistical Study of Literary Vocabulary}.
\newblock Cambridge University Press, 1944.

\bibitem[Miller(1957)]{miller1957}
GA~Miller.
\newblock Some effects of intermittent silence.
\newblock \emph{American Journal of Psychology}, 70:\penalty0 311--314, 1957.

\bibitem[Cancho and Solé(2003)]{ferrericanchoSole2003}
Ramon Ferrer~i Cancho and Ricard~V. Solé.
\newblock Least effort and the origins of scaling in human language.
\newblock \emph{Proceedings of the National Academy of Sciences}, 100\penalty0
  (3):\penalty0 788--791, 2003.

\bibitem[Conrad and Mitzenmacher(2004)]{conradMitzenmacher2004}
B.~Conrad and M.~Mitzenmacher.
\newblock Power laws for monkeys typing randomly: the case of unequal
  probabilities.
\newblock \emph{IEEE Transactions on Information Theory}, 50\penalty0
  (7):\penalty0 1403--1414, 2004.

\bibitem[Piantadosi(2014)]{piantadosi2014}
S.T. Piantadosi.
\newblock Zipf's word frequency law in natural language: A critical review and
  future directions.
\newblock \emph{Psychonomic Bulletin {\&} Review}, 21\penalty0 (5):\penalty0
  1112--1130, 2014.

\bibitem[Norris and McQueen(2008)]{norrisMcQueen2008}
D~Norris and JM~McQueen.
\newblock {{S}hortlist {B}: a {B}ayesian model of continuous speech
  recognition}.
\newblock \emph{Psychol Rev}, 115\penalty0 (2):\penalty0 357--395, Apr 2008.

\bibitem[Balling and Baayen(2012)]{ballingBaayen2012}
LW~Balling and RH~Baayen.
\newblock {{P}robability and surprisal in auditory comprehension of
  morphologically complex words}.
\newblock \emph{Cognition}, 125\penalty0 (1):\penalty0 80--106, Oct 2012.

\bibitem[Hawkins(1994)]{hawkins1994}
JA~Hawkins.
\newblock \emph{A Performance Theory of Order and Constituency}.
\newblock Cambridge University Press, Cambrdige, UK, 1994.

\bibitem[Cohen~Priva(2008)]{cohenPriva2008}
U~Cohen~Priva.
\newblock Using information content to predict phone deletion.
\newblock In \emph{Proceedings of the 27th West Coast Conference on Formal
  Linguistics}, pages 90–--98, Somerville, MA, 2008. Cascadilla Proceedings
  Project.

\bibitem[Levy(2008)]{levy2008}
R~Levy.
\newblock {{E}xpectation-based syntactic comprehension}.
\newblock \emph{Cognition}, 106\penalty0 (3):\penalty0 1126--1177, 2008.

\bibitem[Piantadosi et~al.(2011)Piantadosi, Tily, and
  Gibson]{piantadosiEtAl2011}
ST~Piantadosi, H~Tily, and E~Gibson.
\newblock {Word lengths are optimized for efficient communication}.
\newblock \emph{Proc. Natl. Acad. Sci. U.S.A.}, 108\penalty0 (9):\penalty0
  3526--9, 2011.

\bibitem[Smith and Levy(2013)]{smithLevy2013}
NJ~Smith and R~Levy.
\newblock {{T}he effect of word predictability on reading time is logarithmic}.
\newblock \emph{Cognition}, 128\penalty0 (3):\penalty0 302--319, 2013.

\bibitem[Manning and Sch\"{u}tze(1999)]{manningSchutze1999}
CD~Manning and H~Sch\"{u}tze.
\newblock \emph{Foundations of Statistical Natural Language Processing}.
\newblock MIT Press, Cambridge, MA, USA, 1999.
\newblock ISBN 0-262-13360-1.

\bibitem[Marian et~al.(2012)Marian, Bartolotti, Chabal, and
  Shook]{marianEtAl2012}
V.~Marian, J.~Bartolotti, S.~Chabal, and A.~Shook.
\newblock {{C}{L}{E}{A}{R}{P}{O}{N}{D}: cross-linguistic easy-access resource
  for phonological and orthographic neighborhood densities}.
\newblock \emph{PLoS ONE}, 7\penalty0 (8):\penalty0 e43230, 2012.

\bibitem[Mandelbrot(1954)]{mandelbrot1954SGS}
B~Mandelbrot.
\newblock Simple games of strategy occurring in communication through natural
  languages.
\newblock \emph{Transaction of the {IRE} Professional Group on Information
  Theory {PGIT}}, 3\penalty0 (3):\penalty0 124--137, 1954.

\bibitem[Vitevitch and Luce(1999)]{vitevitchLuce1999}
MS~Vitevitch and PA~Luce.
\newblock Probabilistic phonotactics and neighborhood activation in spoken word
  recognition.
\newblock \emph{J Mem Lang}, 40\penalty0 (3):\penalty0 374--408, 1999.

\bibitem[Luce and Large(2001)]{luceLarge2001}
PA~Luce and NR~Large.
\newblock Phonotactics, density, and entropy in spoken word recognition.
\newblock \emph{Lang Cognitive Proc}, 16\penalty0 (5-6):\penalty0 565--581,
  2001.

\bibitem[Marslen-Wilson and Welsh(1978)]{marslenWilsonWelsh1978}
WD~Marslen-Wilson and A~Welsh.
\newblock Processing interactions and lexical access during word recognition in
  continuous speech.
\newblock \emph{Cognitive {P}sychology}, 10\penalty0 (1):\penalty0 29 -- 63,
  1978.

\bibitem[Marslen-Wilson(1987)]{marslen-Wilson1985}
WD~Marslen-Wilson.
\newblock {{F}unctional parallelism in spoken word-recognition}.
\newblock \emph{Cognition}, 25\penalty0 (1-2):\penalty0 71--102, 1987.

\bibitem[Zwitserlood(1989)]{zwitserlood1989}
P~Zwitserlood.
\newblock {{T}he locus of the effects of sentential-semantic context in
  spoken-word processing}.
\newblock \emph{Cognition}, 32\penalty0 (1):\penalty0 25--64, 1989.

\bibitem[Eberhard et~al.(1995)Eberhard, Spivey-Knowlton, Sedivy, and
  Tanenhaus]{eberhardEtAl1995}
KM~Eberhard, MJ~Spivey-Knowlton, JC~Sedivy, and MK~Tanenhaus.
\newblock {{E}ye movements as a window into real-time spoken language
  comprehension in natural contexts}.
\newblock \emph{J Psycholinguist Res}, 24\penalty0 (6):\penalty0 409--436,
  1995.

\bibitem[Luce and Pisoni(1998)]{lucePisoni1998}
PA~Luce and DB~Pisoni.
\newblock Recognizing spoken words: {T}he neighborhood activation model.
\newblock \emph{Ear Hear}, 19\penalty0 (1):\penalty0 1--36, 1998.

\bibitem[Coltheart et~al.(1977)Coltheart, Davelaar, Jonasson, and
  Besner]{coltheartEtAl1977}
M~Coltheart, E~Davelaar, JT~Jonasson, and D~Besner.
\newblock {Access to the internal lexicon}.
\newblock In S~Dornic, editor, \emph{Attention and Performance VI}, pages
  535--555. Lawrence {E}rlbaum {A}ssociates, 1977.

\bibitem[Genzel and Charniak(2002)]{genzelCharnaik2002}
D~Genzel and E~Charniak.
\newblock {Entropy rate constancy in text}.
\newblock \emph{Proceedings of the 40th Annual Meeting of the Association for
  Computational Linguistics (ACL)}, pages 199--206, 2002.

\bibitem[Aylett and Turk(2004)]{aylettTurk2004}
M~Aylett and A~Turk.
\newblock {{T}he smooth signal redundancy hypothesis: a functional explanation
  for relationships between redundancy, prosodic prominence, and duration in
  spontaneous speech}.
\newblock \emph{Lang Speech}, 47\penalty0 (1):\penalty0 31--56, 2004.

\bibitem[Levy and Jaeger(2007)]{levyJaeger2007}
R~Levy and TF~Jaeger.
\newblock Speakers optimize information density through syntactic reduction.
\newblock In B~Sch\"{o}lkopf, J~Platt, and T~Hoffman, editors, \emph{Advances
  in Neural Information Processing Systems 19}, pages 849--856, Cambridge, MA,
  2007. MIT Press.

\bibitem[Brants and Franz(2009)]{googleEuropeanWeb1T}
T~Brants and A~Franz.
\newblock Web 1{T} 5-gram, 10 {E}uropean {L}anguages {V}ersion 1
  {LDC}2009{T}25, 2009.

\bibitem[Brants and Franz(2006)]{googleEnglishWeb1T}
T~Brants and A~Franz.
\newblock Web 1{T} 5-gram {V}ersion 1 {LDC}2006{T}13, 2006.

\bibitem[Michel et~al.(2011)]{michelEtAl2011}
J~Michel et~al.
\newblock Quantitative analysis of culture using millions of digitized books.
\newblock \emph{Science}, 331\penalty0 (6014):\penalty0 176--182, 2011.
\newblock \doi{10.1126/science.1199644}.

\bibitem[Tiedemann(2012)]{tiedemann2012}
J~Tiedemann.
\newblock Parallel data, tools and interfaces in {OPUS}.
\newblock In \emph{Proceedings of the Eight International Conference on
  Language Resources and Evaluation (LREC'12)}, 2012.

\bibitem[Aylett and Turk(2006)]{aylett2006}
M~Aylett and A~Turk.
\newblock Language redundancy predicts syllabic duration and the spectral
  characteristics of vocalic syllable nuclei.
\newblock \emph{The Journal of the Acoustical Society of America}, 119\penalty0
  (5):\penalty0 3048--3058, 2006.

\bibitem[Bell et~al.(2009)Bell, Brenier, Gregory, Girand, and
  Jurafsky]{bellEtAl2009}
A~Bell, JM~Brenier, M~Gregory, C~Girand, and D~Jurafsky.
\newblock Predictability effects on durations of content and function words in
  conversational english.
\newblock \emph{Journal of Memory and Language}, 60\penalty0 (1):\penalty0
  92--111, 2009.

\bibitem[Gahl et~al.(2012)Gahl, Yao, and Johnson]{gahlEtAl2012}
S~Gahl, Y~Yao, and K~Johnson.
\newblock Why reduce? phonological neighborhood density and phonetic reduction
  in spontaneous speech.
\newblock \emph{Journal of Memory and Language}, 66\penalty0 (4):\penalty0
  789--806, 2012.

\bibitem[oed(2016)]{oed2016}
Oed online, 2016.

\bibitem[Mahowald et~al.(2013)Mahowald, Fedorenko, Piantadosi, and
  Gibson]{mahowaldEtAl2013}
K.~Mahowald, E.~Fedorenko, S.~T. Piantadosi, and E.~Gibson.
\newblock {{I}nfo/information theory: speakers choose shorter words in
  predictive contexts}.
\newblock \emph{Cognition}, 126\penalty0 (2):\penalty0 313--318, Feb 2013.

\bibitem[Surendran and Niyogi(2006)]{surendran2006quantifying}
D~Surendran and P~Niyogi.
\newblock Quantifying the functional load of phonemic oppositions, distinctive
  features, and suprasegmentals.
\newblock \emph{Amsterdam Studies in the Theory and History of Linguistic
  Science Series 4}, 279:\penalty0 43, 2006.

\bibitem[Wedel et~al.(2013)Wedel, Kaplan, and Jackson]{wedel2013high}
A~Wedel, A~Kaplan, and S~Jackson.
\newblock High functional load inhibits phonological contrast loss: A corpus
  study.
\newblock \emph{Cognition}, 128\penalty0 (2):\penalty0 179--186, 2013.

\bibitem[Baayen et~al.(1995)Baayen, Piepenbrock, and Gulikers]{baayenEtAl1995}
HR~Baayen, R~Piepenbrock, and L~Gulikers.
\newblock The {CELEX} lexical database. {R}elease 2 ({CD-ROM}), 1995.

\bibitem[Futrell et~al.(In press)Futrell, Albright, Graph, and
  O'Donnell]{futrellInPress}
R.~Futrell, A.~Albright, P~Graph, and T.J. O'Donnell.
\newblock A generative model of phonotactics.
\newblock \emph{Transactions of the Association for Computational Linguistics},
  In press.

\bibitem[Strand and Liben-Nowell(2016)]{strand2016}
J~Strand and D~Liben-Nowell.
\newblock Making long-distance relationships work: Quantifying lexical
  competition with hidden markov models.
\newblock \emph{Journal of Memory and Language}, 90:\penalty0 88 -- 102, 2016.

\bibitem[Lin et~al.(2012)Lin, Michel, Aiden, Orwant, Brockman, and
  Petrov]{lin2012syntactic}
Yuri Lin, Jean-Baptiste Michel, Erez~Lieberman Aiden, Jon Orwant, Will
  Brockman, and Slav Petrov.
\newblock Syntactic annotations for the google books ngram corpus.
\newblock In \emph{Proceedings of the ACL 2012 system demonstrations}, pages
  169--174. Association for Computational Linguistics, 2012.

\bibitem[Smith(Submitted)]{smithSubmitted}
NJ~Smith.
\newblock {ZS}: A file format for efficiently distributing, using, and
  archiving record-oriented data sets of any size.
\newblock Submitted.

\bibitem[Chen and Goodman(1999)]{chenGoodman1999}
SF~Chen and J~Goodman.
\newblock An empirical study of smoothing techniques for language modeling.
\newblock \emph{Comput Speech \& Lang}, 13\penalty0 (4):\penalty0 359--393,
  1999.

\bibitem[Stolcke(2002)]{stolcke2002}
A~Stolcke.
\newblock Proceedings of icslp.
\newblock volume~2, pages 901--904, 2002.

\end{thebibliography}

\end{document}